\documentclass{article}
\pdfoutput=1

\usepackage[preprint, nonatbib]{neurips_2024}

\usepackage[utf8]{inputenc} 
\usepackage[T1]{fontenc}    
\usepackage{hyperref}       
\usepackage{url}            
\usepackage{booktabs}       
\usepackage{amsfonts}       
\usepackage{nicefrac}       
\usepackage{microtype}      
\usepackage{xcolor}         
\usepackage{graphicx}
\usepackage{amsmath}
\usepackage{amssymb}
\usepackage{subcaption}
\usepackage{algorithm} 
\usepackage{algcompatible}

\title{Varying Manifolds in Diffusion: From Time-varying Geometries to Visual Saliency}

\author{%
  Junhao Chen\\
  Shandong University \\
  \texttt{jhochen@mail.sdu.edu.cn}
   \And
   Manyi Li\thanks{Corresponding author.} \\
   Shandong University \\
   \texttt{manyili@sdu.edu.cn} \\
   \AND
   Zherong Pan \\
   Tencent America \\
   \texttt{zherong.pan.usa@gmail.com} \\
   \And
   Xifeng Gao \\
   Tencent America \\
   \texttt{gxf.xisha@gmail.com} \\
   \And
   Changhe Tu \\
   Shandong University \\
   \texttt{chtu@sdu.edu.cn} \\
}

\begin{document}

\maketitle

\begin{abstract}
Deep generative models learn the data distribution, which is concentrated on a low-dimensional manifold. The geometric analysis of distribution transformation provides a better understanding of data structure and enables a variety of applications. In this paper, we study the geometric properties of the diffusion model, whose forward diffusion process and reverse generation process construct a series of distributions on manifolds which vary over time. Our key contribution is the introduction of generation rate, which corresponds to the local deformation of manifold over time around an image component. We show that the generation rate is highly correlated with intuitive visual properties, such as visual saliency, of the image component. Further, we propose an efficient and differentiable scheme to estimate the generation rate for a given image component over time, giving rise to a generation curve. The differentiable nature of our scheme allows us to control the shape of the generation curve via optimization. Using different loss functions, our generation curve matching algorithm provides a unified framework for a range of image manipulation tasks, including semantic transfer, object removal, saliency manipulation, image blending, etc. We conduct comprehensive analytical evaluations to support our findings and evaluate our framework on various manipulation tasks. The results show that our method consistently leads to better manipulation results, compared to recent baselines.
\end{abstract}

\section{Introduction}
Modern deep learning architectures rely on a fundamental principle: high-dimensional data resides on a low-dimensional manifold. This principle of dimensionality reduction underlies many recent advancements in generative models, such as Variational Autoencoders (VAEs)~\cite{kingma2022autoencoding} and Generative Adversarial Networks (GANs)~\cite{goodfellow2014generative}, which capture the essential structure of data by learning efficient, low-dimensional representations. Building on these low-dimensional embeddings, a representative approach to learning data distributions involves constructing transformations between the data distribution and a Gaussian distribution. This method allows for the representation of any probability distribution while introducing a geometric mapping that enables further understanding of data structure. For instance, the analysis of the geometric properties of both the data and latent manifolds in GANs enables applications like geodesic interpolation~\cite{Arvanitidis2017LatentSO,shao2017riemannian}.

Most recently, diffusion models~\cite{sohldickstein2015deep,ho2020denoising} have significantly improved the expressivity of generative models, more effectively enabling a single model to represent diverse internet images in the wild. Since then, the analysis of data distributions learned by diffusion models has garnered much attention from researchers. However, this analysis is more challenging than for prior models due to the transformation between the data distribution and Gaussian distribution involving the entire forward diffusion process, leading to time-varying geometric mappings.

In this paper, we propose a metric to gauge the rate of change in the manifold as a function of time. As our key observation, we show that such a change rate corresponds to the rate of information removal during the diffusion process, or the rate of information injection during the reverse generation process. Therefore, we call our metric in the reverse process the \textit{"generation rate"}, which changes over time to define the \textit{"generation curve"}. Further, we propose an efficient and differentiable scheme to approximate the generation rate in the observation space (image space for 2D diffusion models). Utilizing its differentiable nature, we can then manipulate the shape of the generation curve by stochastic optimization, i.e., to match the shape of the curve with a given reference curve. Finally, we show that such optimization corresponds to a unified framework for a range of image manipulation tasks, such as semantic transfer, object removal, saliency manipulation, and image blending. By comprehensively evaluating our approach in all these tasks, we confirm that our framework consistently outperform existing state-of-the-art models. 
\section{Background}
\label{sec:background}
\subsection{Diffusion Process as Stochastic Differential Equations}
The diffusion model~\cite{sohldickstein2015deep, ho2020denoising} is a type of stochastic generative model that gradually adds noise to the original data in a forward diffusion process and generates realistic data samples via a reverse denoising process. It can be formulated as stochastic differential equations (SDEs)~\cite{song2021scorebased} with a continuous time variable $t \in [0,T]$. The forward diffusion process, which evolves a probabilistic distribution towards a more uniform or stable state over time through random perturbations, is written as:
\begin{equation}
dX_t = \mu(X_t, t) \, dt + \sigma(X_t, t) \, dW_t,
\label{eq:sde}
\end{equation}
where $X_t$ represents the state of the process at time $t$, $\mu(X_t, t)$ is the drift coefficient, $\sigma(X_t, t)$ is the volatility coefficient, and $dW_t$ is the differential of a Wiener process.

The reverse SDE, used for denoising and generating data, is formulated as:
\begin{equation}
dX_t = [\mu(X_t, t) - \sigma^2(X_t, t) \nabla_x \log p_t(X_t)] \, dt + \sigma(X_t, t) \, dW_t,
\end{equation}
where $\nabla_x \log p_t(X_t)$ is the score function of the probability density function $p_t(X_t)$. 

We can further derive a deterministic process with trajectories that share the same marginal probability densities as the SDE (Eq.~\ref{eq:sde}). This is formulated as an ordinary differential equation (ODE)~\cite{song2021scorebased}:
\begin{equation}
dX_t = [\mu(X_t, t) - \frac{1}{2}\sigma^2(X_t, t) \nabla_x \log p_t(X_t)] \, dt,
\label{eq:ode}
\end{equation}
In this paper, we adopt this deterministic approach and use its specific discrete form from~\cite{song2020denoising}:
\begin{equation}
\frac {X_{t-\Delta t}}{\sqrt{\alpha_{t-\Delta t}}} = \frac{X_t}{\sqrt{\alpha_t}} + \left( \sqrt{\frac{1 - \alpha_{t-\Delta t}}{\alpha_{t-\Delta t}}} - \sqrt{\frac{1 - \alpha_t}{\alpha_t}} \right) \epsilon^t_\theta(X_t),
\label{eq:ddim_ode}
\end{equation}
where $\alpha_t$ is a time-dependent variable as defined in~\cite{song2020denoising} and $\epsilon^t_\theta(X_t)$ is a neural network with parameter $\theta$ trained to approximate the score function $\nabla_x \log p_t(X_t)$. In this form, one can directly obtain a predicted $\hat{X}_0$ from the linear approximation of $X_t$ by:
\(
\hat{X}_0(X_t) = (X_t - \sqrt{1-\alpha_t}\epsilon^t_\theta(X_t))/\sqrt{\alpha_t},
\label{pred_X0}
\)
which can further be used as an estimation of the generation state.

\subsection{\label{sec:geometric_analysis}Geometric Analysis of Data Manifold}
It is widely accepted that the distribution of high-dimensional observed data, i.e. the distribution of open-domain images, resides on a low-dimensional manifold $M$ embedded in the high-dimensional Euclidean space $R^d$~\cite{Tenenbaum2000AGG,Roweis2000NonlinearDR}.
From this perspective, generative models such as VAE~\cite{kingma2022autoencoding} and GAN~\cite{goodfellow2014generative} learn the mapping from a simple latent space $Z$, usually a Gaussian distribution, to the data manifold $M$. Geometric analysis of these manifold mappings~\cite{Arvanitidis2017LatentSO, shao2017riemannian} provides a valuable tool for studying the structures of both data and latent manifolds, enabling various applications~\cite{Arvanitidis2017LatentSO, shao2017riemannian, chen2018metrics, Arvanitidis2017LatentSO}.

The tangent space $T_{x}M$ computed on manifolds is crucial for local analysis~\cite{Rifai2011ContractiveAE, NIPS2011_d1f44e2f, kumar2017semisupervised}. It has been shown that using local contraction as a penalty during the training of auto-encoders results in better latent representations that are locally invariant to perturbations in the noise directions of the raw input~\cite{Rifai2011ContractiveAE}, through which the tangent space is approximately spanned by the subset of singular vectors corresponding to largest singular values from encoding mapping (see e.g.~\cite{bengio2014representation}). Moreover, in GANs, the open set formed by latent variables ensures that the Jacobian of the generator directly spans the tangent space~\cite{kumar2017semisupervised, shao2017riemannian}. Analyzing directional contraction in such mappings enables the unsupervised discovery of semantically disentangled image editing directions~\cite{ramesh2019spectral, Voynov2020UnsupervisedDO}.

\begin{figure*}[t]
\centering
\includegraphics[width=1\linewidth]{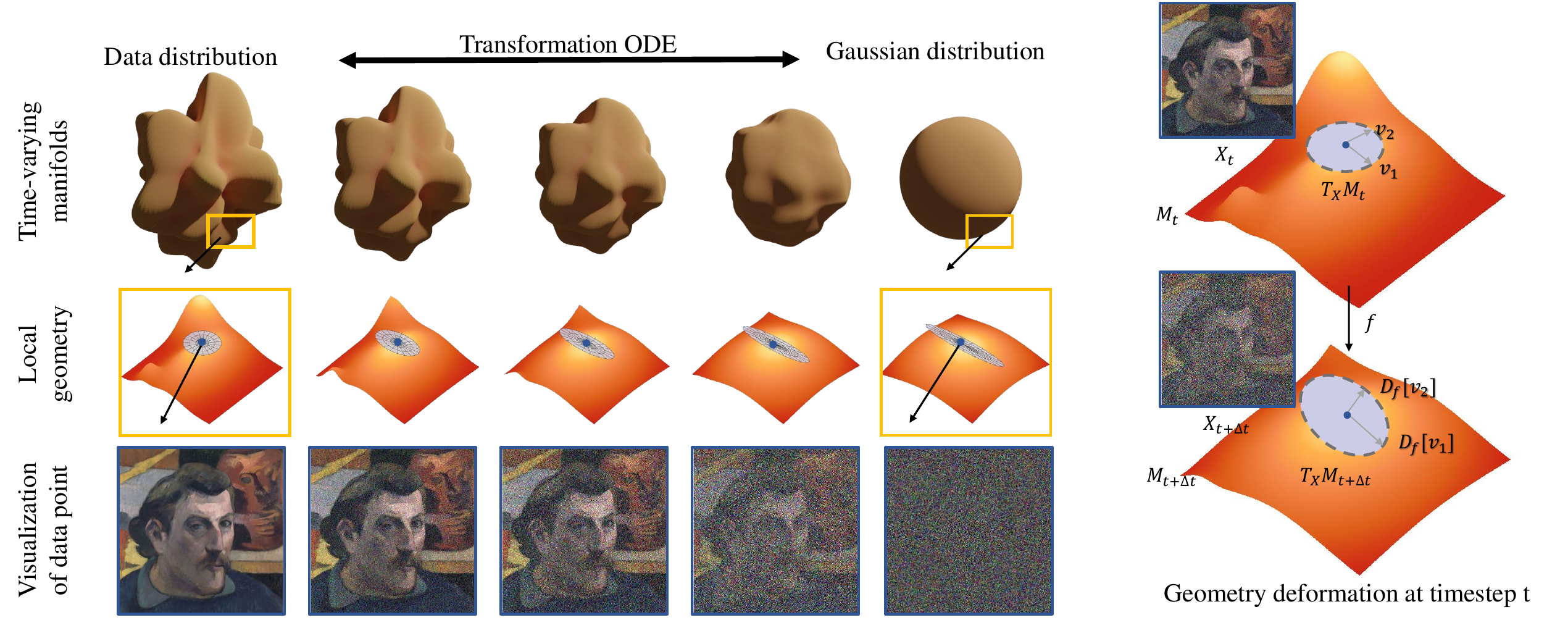}
\caption{The time-varying manifolds in diffusion process and the local geometric transition.}
\label{fig:forward}
\end{figure*}

The generative process of diffusion models can be seen as iteratively modifying a manifold to approximate the true data manifold. This process can be modeled as a series of manifolds denoted as $\{M_t\}$, as shown in Figure~\ref{fig:forward}. Regretfully, for diffusion transformations built on the Euclidean observation space, unlike the straightforward mappings available in VAEs and GANs, the manifolds $\{M_t\}$ lack a direct mapping to a compact latent space. This absence complicates the geometric analysis of the manifold $M_t$. Fortunately, for diffusion models with a U-Net architecture, prior works~\cite{kwon2023diffusion, park2023understanding} have empirically discovered that the encoder layers of the score predictor U-Net can be interpreted as a mapping, denoted as $h_t$, from the transient manifold to a (transient) compact latent space $H_t$. In practice, we can use the power method~\cite{Haas2023DiscoveringID,park2023understanding} to approximate the leading right singular vectors of the Jacobian matrix \( J_{h_t} \) of \( h_t \) that span the tangent space $T_{x}M_t$.

\section{Generation Rate \& Generation Curve}
In this section, we first define our notation of generation rates and generation curves (Section~\ref{sec:define_curve}), and then propose an efficient scheme to approximately compute these curves (Section~\ref{sec:compute_curve}). Finally, we show that these curves correspond to the rate at which the diffusion model generates visual contents, and the fluctuation exhibits high correlation with the visual saliency (Section~\ref{sec:analysis}).

\subsection{Definition}
\label{sec:define_curve}
The key idea in our analysis lies in the temporally local analysis of the time-varying manifolds $\{M_t\}$. 
Between these consecutive manifolds, we define the mapping corresponding to the forward diffusion process as $X_t = f_t(X_{t-\Delta t})$, and the mapping corresponding to the reverse diffusion process as $X_{t-\Delta t} = f_t^{-1}(X_t)$.
Instead of analyzing the mapping $f$ from the latent space to the data manifold, as has been done for GANs and VAEs, we analyze the local geometric distortion between two temporally consecutive manifolds. 
Following a similar reasoning as prior works~\cite{Rifai2011ContractiveAE, bengio2014representation}, we represent the distortion through the deformation between tangent spaces, which can be derived from the Jacobian matrix $J_f$ of the mapping $f$. Specifically, the scaling of a tangent vector is represented by its corresponding singular value.
An important empirical finding is that, under the parametrization in Eq.~\ref{eq:ddim_ode}, the singular values of $J_f$ for tangent vectors almost always fall in the range $(0,1)$, implying that $f_t$ is a contracting mapping when restricted to $M_t$. This is consistent with the fact that the diffusion process removes information from the data and injects entropy into the distribution on $M_t$ as $t$ increases. Similarly, we can consider the reverse process and the associated map $f_t^{-1}$, whose Jacobian $J_{f^{-1}}$ empirically has singular values in the range $(1, \infty)$. This corresponds to the process of injecting information into the distribution by reducing its entropy. 
Using these observations, we define our information generation rate as the norm of the directional derivative:
\begin{align}
\label{eq:generation_rate}
D_{f_t^{-1}}(X_t)[v]: T_xM_t \mapsto T_xM_{t-\Delta t},
\end{align}
along a tangent-space variation $v$ in the observation space.
Intuitively, for two images noised to the level $t$ and separated by $v$ in the observation space, they are separated by $\|D_{f_t^{-1}}(X_t)[v]\|$ in the previous noised level $t-\Delta t$. The variation vector $v$ provides us with an important extra degree of freedom, allowing us to investigate the generation rate in a subset of the observation space. For example, by setting $v$ to correspond to a specific image component, we can observe the generation rate for that component. The selection of $v$ is crucial to our various image manipulation tasks.

A potential pitfall of Eq.~\ref{eq:generation_rate} lies in the requirement that $v$ is a tangent-space variation. In practice, however, 
an arbitrarily sampled variation $v\in R^d$ to a noised image $X_t$ might not lie in the tangent space $T_{x}M_t$. To mitigate this flaw, we define the projection operator $\text{Proj}(v)$ as the projection of $v$ into the subspace of the first $K$ leading singular vectors. Combining these definitions, we define the projected generation rate:
\begin{equation}
r_t(X_t,v)\triangleq \|D_{f_t^{-1}}(X_t)[\text{Proj}(v)]\|,
\end{equation}
which allows an arbitrary variation $v$ in the ambient space to be used. We then define the generation curve $c(X_T,v)$ as the curve of $r_t(X_t,v)$ calculated over all discrete time instances during the entire generation process for a fixed variation $v$, starting from $X_T$ for $t \in [0,T]$. Assuming the fixed path $\{X_0,X_1,...,X_T\}$ derived from Eq.~\ref{eq:ddim_ode}, any variable $X_t$ corresponds to the same curve. Therefore, we do not distinguish between the input variables, i.e. $c(X_T,v)\triangleq c(X_0,v)$.
\begin{figure}[t]
\centering
\includegraphics[width=1\linewidth]{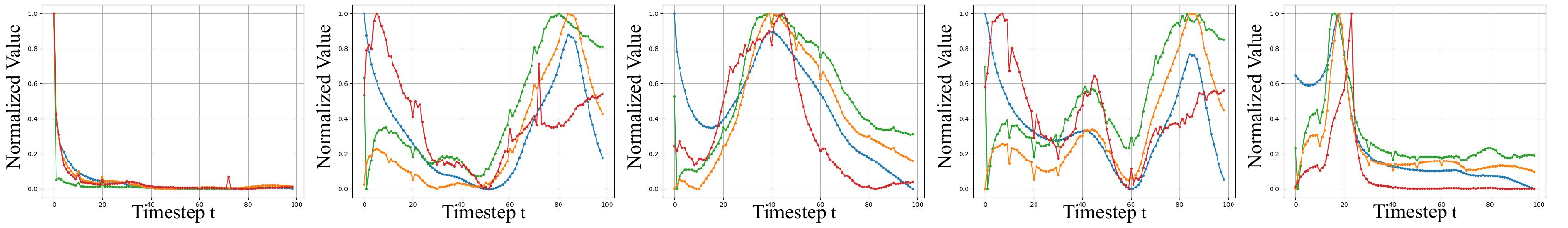}
\caption{The generation curves approximated by four different methods: $\|D_{f^{-1}}(X_t)[\text{Proj}(v)]\|$ \textcolor{green}{\textbullet}, $\|\nabla_x^2 \log p_t(X_t)\cdot\text{Proj}(v)\|$ \textcolor{red}{\textbullet}, $\|D_{h_t}(X_t)[\text{Proj}(v)]\|$ \textcolor{orange}{\textbullet}, $\|\|D_{h_t}(X_t)[v]\|\|$ \textcolor{blue}{\textbullet}, normalized into range $[0,1]$.}
\label{fig:curves}
\end{figure}
\subsection{Approximate Computation of Generation Curve}
\label{sec:compute_curve}
Regretfully, it is challenging to further apply \( c(X_T,v) \) in applications as the projection operation \(\text{Proj}(v)\) requires the calculation of the tangent space at each moment. In the context of using the leading singular vectors of \( J_h \) as the tangent basis, the need for Singular Value Decomposition (SVD) often makes it  non-differentiable, especially when applying power methods for high-dimensional data.
In this section, we propose a practical approximation scheme for the computation of the generation rate $r_t(X_t,v)$. We start by rewriting the generation rate based on Eq.~\ref{eq:ddim_ode}:
\begin{equation}
D_{f^{-1}}(X_t) = A(t)I + B(t)\nabla_x^2 \log p_t(X_t),
\label{eq:Differential of reverse diffusion}
\end{equation}
where \( A(t) \) and \( B(t) \) are time-dependent terms, \( I \) is the identity matrix, and \( \nabla_x^2 \log p_t(X_t) \) is the Hessian of \( \log p_t(X_t) \). We note that the content of the noised image is contained in the score function, so we discard other terms and approximate:
\begin{equation}
\|D_{f^{-1}}(X_t)[\text{Proj}(v)]\|\approx\|\nabla_x^2 \log p_t(X_t)\cdot\text{Proj}(v)\|.
\end{equation}
Next, we utilize the fact that the output space of mapping \( h_t \) is the bottleneck for the score-function predicting network $\epsilon^t_\theta(X_t)$. Thus, we propose replacing the differential of $\epsilon^t_\theta(X_t)$ with that of \( h \):
\begin{equation}
\|\nabla_x^2 \log p_t(X_t)\cdot\text{Proj}(v)\|\approx \|D_{h_t}(X_t)[\text{Proj}(v)]\|.
\end{equation}
Given that \( H_t \) itself is a compact latent space, as introduced in Section~\ref{sec:background}, the forward differential \( D_{h_t}(X_t)[v] \) of any vector \( v \) closely approximates the tangent space estimated by \( h_t \).
Therefore, we omit the projection operator and approximate the generation rate by:
\begin{equation}
\begin{aligned}
r_t(X_t,v)=&\|D_{f_t^{-1}}(X_t)[\text{Proj}(v)]\|\approx\|\nabla_x^2 \log p_t(X_t)\cdot\text{Proj}(v)\|
\\
\approx&\|D_{h_t}(X_t)[\text{Proj}(v)]\|\approx\|D_{h_t}(X_t)[v]\|.
\end{aligned}
\label{eq:curve_computation}
\end{equation}
In Figure~\ref{fig:curves}, we plot the generation curve with $r_t(X_t,v)$ approximated by four different methods. As expected, the four curves exhibit a similar trend, with critical points appearing at comparable noise levels (diffusion time). Although the exact values of these curves differ due to variations in space, 
their trend is sufficient for further analysis and image manipulation tasks. Since the evaluation of the last approximation $\|D_{h_t}(X_t)[v]\|$ is the most computationally efficient and inherently differentiable, we adopt this approximation scheme in our subsequent analysis and applications.
\subsection{\label{sec:analysis}Connection to Visual Saliency}

\begin{figure}
    \centering
    \includegraphics[width=1\linewidth]{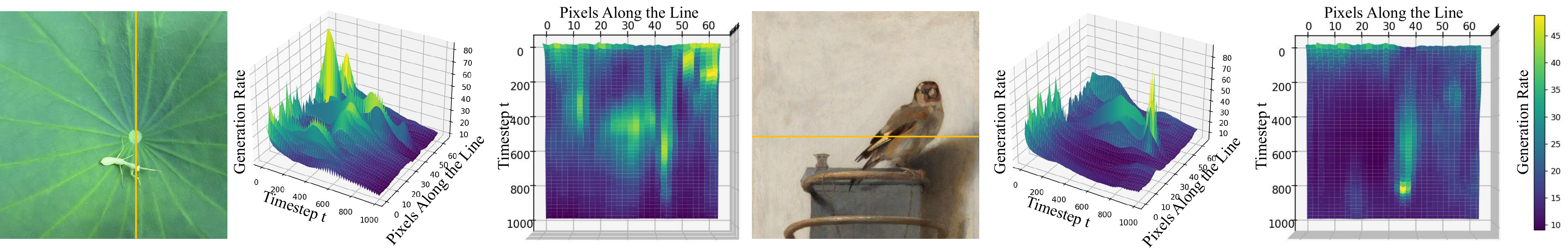}
    \caption{Generation curves for a column or row of image pixels (yellow). The generation curves fluctuate significantly at the pixels with high visual saliency, such as the wing tip of the bird.}
    \label{fig:curve_ana}
\end{figure}

\begin{figure}
    \centering
    \includegraphics[width=1\linewidth]{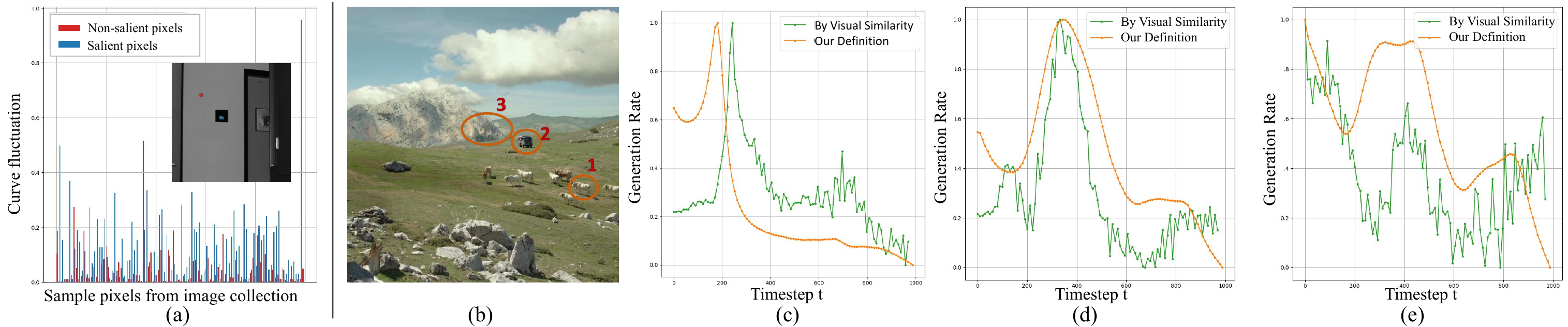}
    \caption{Left: visual saliency analysis with (a) curve fluctuation statistics of salient (blue) and non-salient (red) pixels as well as an example image. Right: generation curves at spatial locations on the image (b). (c), (d) and (e) show the perceptual-based (green) and our generation curve (orange) corresponding to the specified three regions.}
    \label{fig:visual_experiments}
\end{figure}

In this section, we present our comprehensive analysis showing that the generation curve is strongly connected to visual properties on images. To this end, we set the vector $v$ as a unit vector $v=e_{ij}$ that takes the value $1$ at $ij$-th pixel and zero otherwise. This setting allows us to investigate the rate of information generation for a single channel of a pixel.
Considering the additive generation rate, we define the generation rate of a pixel as the average of its internal channels. In implementation, we take the $0$-th channel to represent a pixel for the acceleration purpose, due to the characteristics of latent diffusion model (LDM)~\cite{rombach2022highresolution} as described in Appendix~\ref{sec:visual_analysis_appendix}.

In Figure~\ref{fig:curve_ana}, we plot the generation curve for a column or row of image pixels. We noticed that pixels with high visual saliency, such as the wing tip and the body of the bird, the generation curve fluctuates significantly. At other locations, the curve becomes much smoother except for a sharp rise when $t$ approaches 0. We further conduct a large-scale analysis by using 100 pictures from the visual saliency dataset~\cite{borji2015cat2000}. For each image, we sample one salient pixel and one non-salient pixel from the ground truth. For both pixels, 
we take the windowed variance to measure the fluctuation of its generation curve, and the statistics are shown in figure~\ref{fig:visual_experiments}. 
(Refer to Appendix~\ref{app:saliency_val} for more details.)
For $86\%$ of the images, higher visual saliency leads to higher fluctuation, validating the consistent correspondence between curve fluctuation and visual saliency.

For other morphological factors of the curve beyond its fluctuation, such as the position and curvature of the peaks, we experimentally found that these are determined by more specific and low-level visual properties of the object. For instance, image patches with different materials like grass and ground exhibit different curve shapes even though they share similar visual saliency. 

We further notice an alternative way of defining the generation rate, which is inspired by the generation state derived from~\cite{song2020denoising}. 
For each noised image $X_t$, we can compare the visual similarity of its predicted $\hat{X}_0$ to the real image $X_0$, e.g., by computing the perceptual distance~\cite{zhang2018unreasonable, caron2021emerging} between the two images. Such computation can be performed with respect to an image patch by applying a region mask. We consider this similarity as the generation state, and the time derivative of this state can also be interpreted as the generation rate. In Figure~\ref{fig:visual_experiments}, we compare the generation curve computed using our definition and this alternative definition for objects (details in Appendix~\ref{app:generation_val}). We notice that the two curves exhibit similar trends while our curve has much less noise.
Besides, due to the nature of visual metrics, this method is applicable only to regions with semantically complete objects, and often exhibits heavy noise or even negative generation rates in areas with less prominent visual features, especially with background patches. These limitations prevent this alternative method from providing reliable indications of the information generation rate, limiting its usage in assisting other tasks, such as observing the generation process of objects, or selecting an editing timestep~\cite{kwon2023diffusion,choi2022perception}.

\section{\label{sec:curve_matching}Curve Matching}
We utilize the differentiable scheme of our generation rate to define an optimization procedure applicable to various image manipulation tasks. Since the generation curve is intrinsically related to certain visual properties, it enables us to manipulate those properties by modifying the curves.  For simplicity, we take single-channel images to describe our algorithm. In practice, we enumerate all the channels of the images and invoke our algorithm equivalently. 

The proposed curve matching algorithm manipulates the visual properties of an image patch by aligning the shapes of curves. Its input contains an image $X_0$, a specified reference generation curve denoted as $c^\star$ with the corresponding reference generation rates $r_t^\star$, and a source image patch $\mathcal{A}$ to be edited. We can define the marginalized generation curve over the image patch as: 
\begin{align*}
c(X_0,\mathcal{A})\triangleq\sum_{e_{ij}\in\mathcal{A}}c(X_0,e_{ij})l(e_{ij}),
\end{align*}
where $l(e_{ij})$ is a uniform distribution over the pixels in the patch $\mathcal{A}$. Our goal is to search for a desired image $\bar{X}_0$ whose marginalized generation curve matches that of $c^\star$, formulated as:
\begin{align}
\label{eq:opt}
\text{argmin}_{\bar{X}_0}D(c(X_0,\mathcal{A})|c^\star),
\end{align}
where $D$ is the distance metric. Since it is challenging to normalize the generation curve over the image patch, i.e. $c(X_t,\mathcal{A})$, we use the total variance (TV) distance as the metric $D$. 

However, directly optimizing $X_0$ in the ambient space often leads to distortions due to the restricted nature of the data distribution. Accordingly, based on Eq.~\ref{eq:ddim_ode}, we select its corresponding $X_t$ as the optimization variable, which is situated in a diffused distribution and is thus more fault-tolerant.
On the other hand, due to the computational burden, we conduct local optimization by sampling a pixel and a timestep per iteration.
In summary, for each iteration, we sample a pixel $e_{ij}$ and a time $t_k$ based on the pixel distribution $l(e_{ij})$ and the generation rate distribution $c(X_0,\mathcal{A})$, respectively. The optimization objective is thus rewritten as:
\begin{align}
\label{eq:opt_p}
\text{argmin}_{X_t}E_{c(t_s|X_t,\mathcal{A}),l(e_{ij})}|c(t_s|X_t,e_{ij})-c^\star(t_s)|,
\end{align}
where $|\cdot|$ is the $L_1$ norm.  We then optimize $X_t$ by $X_t\gets X_t-\eta\nabla_{X_t}|r_{t_s}(X_{t_s},e_{ij})-r_{t_s}^\star|$ with $\eta$ as the learning rate using the SGD optimizier. After the optimization, we recover $\bar{X}_0$ from the optimized $X_t$ by Eq.~\ref{eq:ddim_ode}.
The optimization algorithm can be applied to a range of image manipulation tasks by specifying the reference generation curve and adding additional constraints in the objective function.

\paragraph{Localized Modification} A typical requirement in applications is to edit only the image inside a given patch $\mathcal{A}$. However, our standard SGD scheme often modifies the entire image, which is undesirable. 
Simply applying a mask to $X_t$ does not solve the problem either, because the mask is applied at the noised level, while the corresponding noiseless image $\bar{X}_0$ can still contain modifications outside $\mathcal{A}$. Instead, we propose to iteratively blend the noised original image, $X_t$, and the noised optimized image, $\bar{X}_t$, via
$X_t\gets \hat{X}_t\odot\mathcal{A}+X_t\odot\bar{\mathcal{A}}$, where $\bar{\mathcal{A}}$ is the complement of $\mathcal{A}$ and $\odot$ is the pixel-wise product. We perform such blending for $t>t_\text{blend}$ every $70$ iterations. Here we only apply blending at a sufficiently high noise level ($t>t_\text{blend}$) so that the noiseless image has seamless patch boundaries around $\mathcal{A}$ due to the nature of the reverse generation process.

\section{Image Manipulation Applications}
Many image manipulation tasks can be considered as the transformation of visual properties. This section demonstrates how our curve matching algorithm can flexibly perform various image manipulations using only a pre-trained unconditional diffusion model. Our generation curve provides a unified framework for many tasks and eliminates the need to train separate models for each of them or require large datasets for specific domains.
For more results, please refer to our Appendix~\ref{app:application}.

\begin{figure}
    \centering
    \includegraphics[width=1\linewidth]{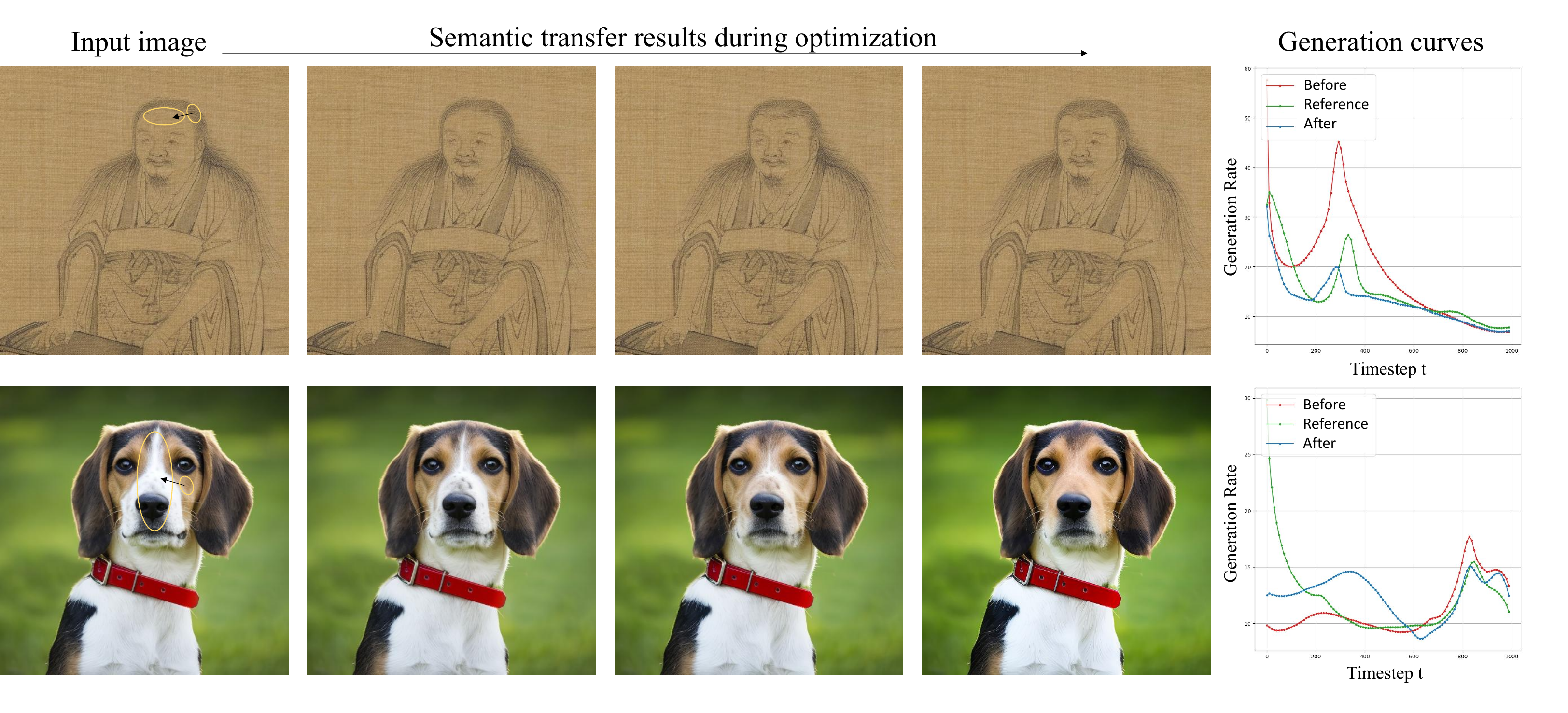}
    \caption{Semantic transfer results. Left: the input image and the transfer results during optimization. Right: generation curves before and after optimization, as well as the reference curve.}
    \label{fig:visual_transfer}
\end{figure}
\begin{figure}
    \centering
    \includegraphics[width=1\linewidth]{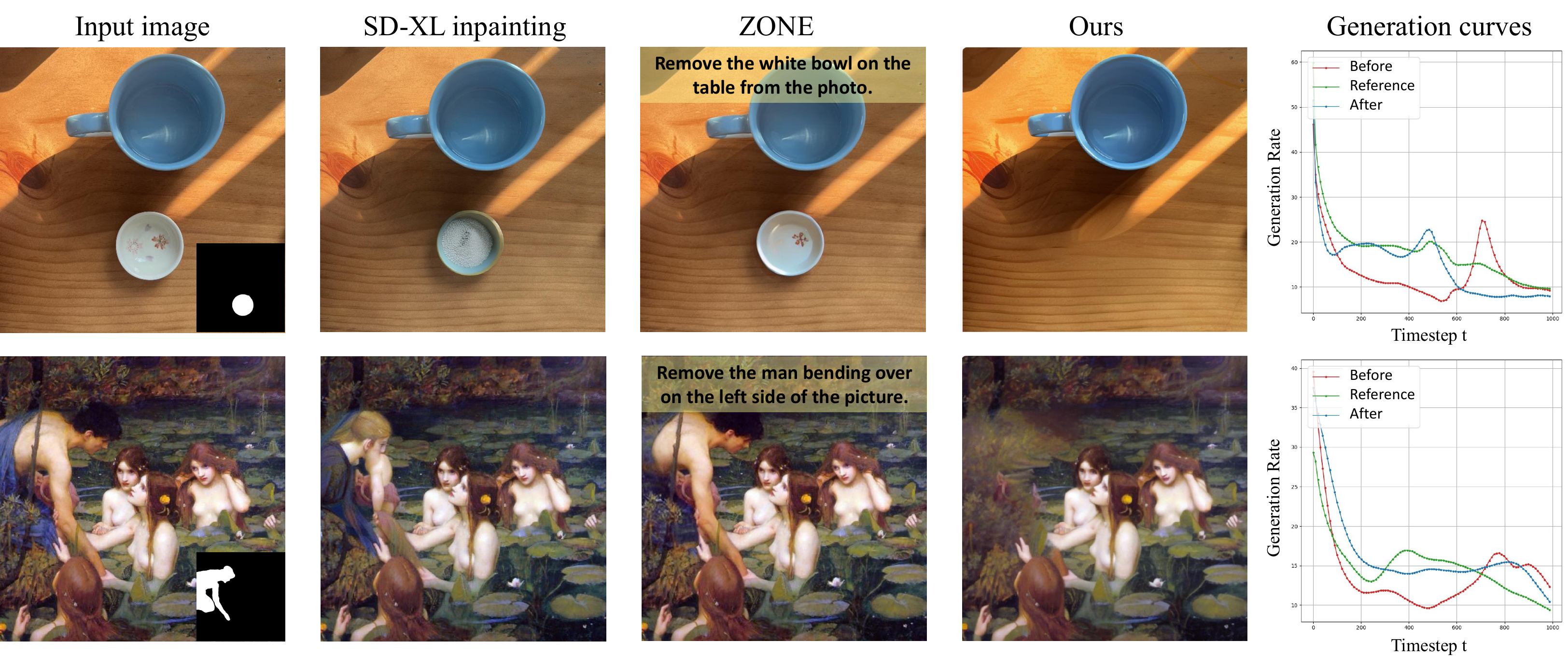}
    \caption{Object removal comparison. From left to right: the input image and object mask, the results of alternative approaches~\cite{podell2023sdxl, li2024zone} and ours, corresponding generation curves during optimization.}
    \label{fig:removal_visual}
\end{figure}

\subsection{Semantic Transfer}
The semantic transfer task modifies a source region to match the semantic properties (e.g., color, material, texture) of a reference while maintaining other properties (e.g., depth, shape) unchanged.
Due to the set of visual properties coupled within the curve, this problem can be inherently dealt with using our curve matching algorithm by specifying the reference curve of a pixel with desired semantic properties.
To avoid meaningless transfers from arbitrary references, we select the reference pixel from a region adjacent to the target area, limiting their difference within the expected properties.

Figure~\ref{fig:visual_transfer} illustrates our semantic transfer results as well as the generation rate curves before and after the optimization. In the second row, the goal is to turn the white fur on the dog's face to a brown color. We select the reference pixel on its cheek, which has the desired brown color and is similar to the source area in terms of other aspects, e.g., both areas are fur on the dog. After optimization, the optimized curves align with the specified reference, and the white fur gradually turns brown.

\subsection{Object Removal}
Object removal involves replacing an object with the background it obscured, while keeping the rest of the image unchanged.  
Our curve matching algorithm addresses this by transferring the visual properties of the background to the pixels of the object to be removed. This process shares the same pipeline as semantic transfer, with the reference pixel selected from the expected background. 
\begin{table}[ht]
    \centering
    \begin{tabular}{|c|c|c|c|}
        \hline
        Method & CLIP$_{dir}$↑& CLIP$_{sim}$↓ & $DINO_{sim}$↓\\
        \hline
        ZONE~\cite{li2024zone} & 0.2589 & 0.4824 & 0.433\\
        SD-XL inpainting~\cite{podell2023sdxl} & 0.2617&  0.4787 & 0.422 \\
        Ours & \textbf{0.2629} & \textbf{0.4782} & \textbf{0.416}\\
        \hline
    \end{tabular}
    \caption{Quantitative comparison of methods}
    \label{tab:removal_quantitative1}
\end{table}

Figure~\ref{fig:removal_visual} compares the object removal results of our method and two recent approaches, i.e., SD-XL inpainting~\cite{podell2023sdxl} and an instruction-based method, called ZONE~\cite{li2024zone}.  
The shortcomings of these two types of methods are primarily in the following aspects: Inpainting methods that are trained on masked images can be considered a form of re-sampling from the true image distribution conditioned on the unmasked region~\cite{rout2023theoretical}. Although such re-sampling from high-density regions sometimes corresponds to the background, thereby achieving the goal of object removal, this approach is not stable. Occasionally, the object to be removed may be covered by another object. 
For instruction-based methods, they achieve image editing through a pre-trained model that accepts textual instructions. However, they sometimes fail to identify the objects described in the text instruction. In the case of removal, they often fail when dealing with complex occlusion scenarios.  

We evaluate the performance on the test set from Emu Edit benchmark~\cite{sheynin2023emu}, where the input images, image captions before and after object removal, and the text instructions are provided. For the additional inputs required by our method, such as the mask and reference point, we use a segmentation tool and select the spatial locations. 
Since there is no exact metric for object removal tasks, we select three similarity-based metrics with CLIP~\cite{radford2021learning} and DINO~\cite{caron2021emerging}, i.e., CLIP$_{dir}$ for the similarity between the caption encoding difference and the image encoding difference before and after removal~\cite{sheynin2023emu},  CLIP$_{sim}$ and DINO$_{sim}$ for the similarity between the editing area before and after removal. The quantitative results reported in Table~\ref{tab:removal_quantitative1} validates the superior performance of our approach.

\subsection{Saliency Manipulation}
\begin{figure}
    \centering
    \includegraphics[width=1\linewidth]{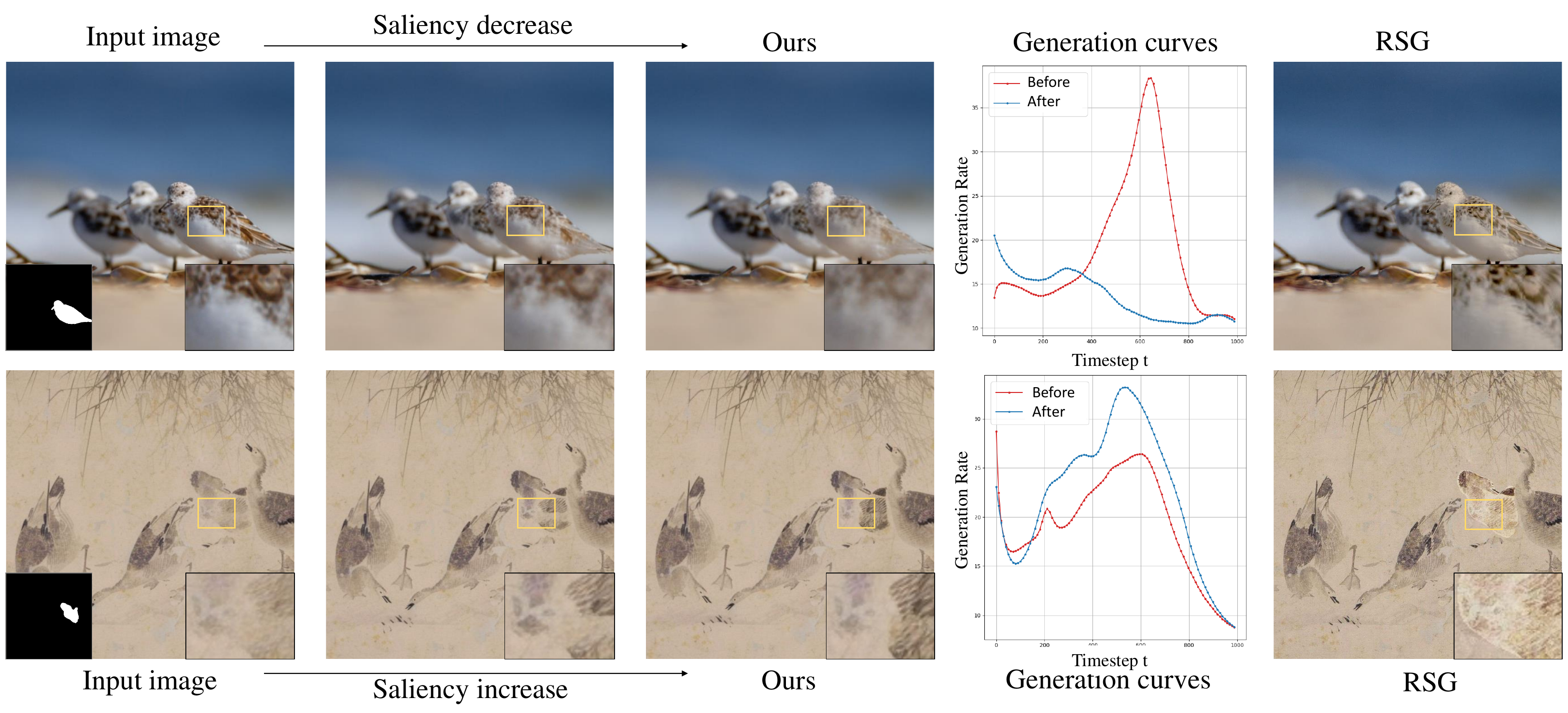}
    \caption{Saliency manipulation comparison. From left to right: input images and masks, intermediate and final results of ours, generation curves during optimization, results of an alternative approach~\cite{miangoleh2023realistic}.}
    \label{fig:saliency_editing_visual}
\end{figure}
Saliency manipulation involves altering the saliency of an object while maintaining its identity. The high correlation between visual saliency and the fluctuation of the generation curve allows us to adjust saliency by modifying these fluctuations during curve optimization. Directly specifying a reference curve pattern with only a different fluctuation is challenging. However, we observe that, except for a brief period when $t$ approaches 0, i.e., $t<t_b$, 
the salient curve is consistently higher and has distinct peaks compared to the non-salient curve, which maintains a lower and fixed value.
Therefore, to increase (resp. decrease) saliency, we simply maximize (minimize) generation rates for $t>t_b$, with $t_b=200$ in our implementation, where the reference curve is implicitly specified. We also use the feature alignment loss from UNet as described in~\cite{mou2023dragondiffusion} to preserve image content.

We present the saliency editing results in Figure~\ref{fig:saliency_editing_visual}, comparing our results and those of a recent approach RSG~\cite{miangoleh2023realistic}. The results indicate that using curves as a measure and quantification of saliency is reasonable. By contrast, it is challenging for the existing methods to learn accurate visual saliency from the eye-tracking data. Consequently, they often focus on patterns such as color contrast and brightness, thus failing to produce harmonized results across diverse natural images.

\subsection{Image Blending}
\begin{figure}
    \centering
    \includegraphics[width=1\linewidth]{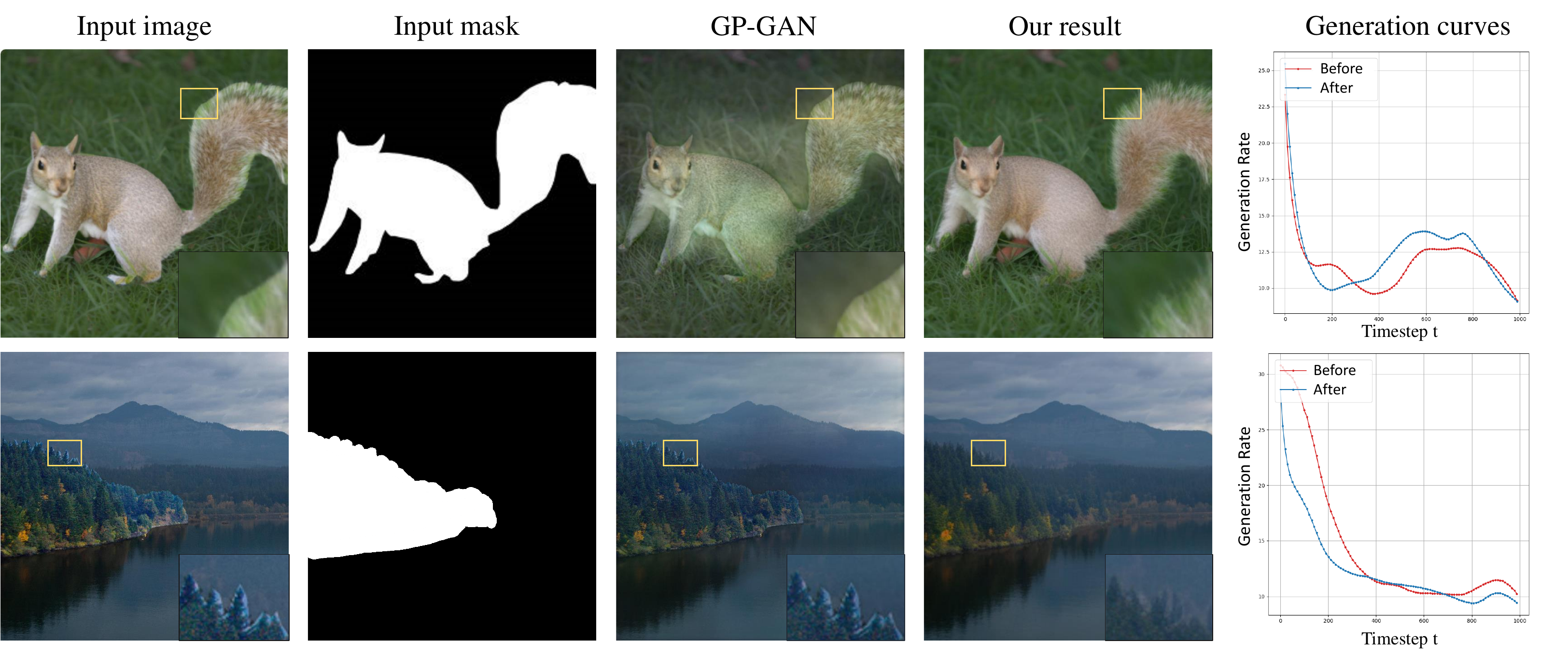}
    \caption{Image blending comparison. From left to right: input image, input mask, results of GP-GAN~\cite{wu2019gpgan}, our results, and the corresponding generation curves.}
    \label{fig:blending_visual}
\end{figure}
The Image blending task aims at blending a foreground image with a background image on boundaries. To accomplish this task, given a composite image, we define the boundary between the foreground and background as the salient region, since the boundaries contain undesirable and eye-catching seams. Therefore, we propose to reduce the visual saliency of the boundary region to make a natural transition. 
With our curve matching algorithm, we follow the same way as our saliency manipulation application, but minimize only the visual saliency at the boundary region of the foreground object. 

In Figure~\ref{fig:blending_visual}, we show the results of our method and an existing approach~\cite{wu2019gpgan}. In our experiments, we found that existing methods often perform well on specific types of images, and fail to produce universally satisfactory results across a variety of natural images. 
Instead, our approach consistently produces visually pleasing boundaries for the composite images.
\section{Conclusion}

In this work, we propose the generation rate, which corresponds to the geometric deformation of the manifold over time around a image component. Through comprehensive analytical evaluations, we show that the time-varying geometric deformation exhibits a high correlation with visual saliency of the image component. In addition, manipulating the generation curves with different loss functions provides a unified framework for a row of image manipulation tasks. 
Future research could explore more applications and address the limitations of our generation curve.
For instance, our curve optimization algorithm requires first-order differentiation computation and thus requires approximately 10 minutes for 300 iterations including the pre-processing, running on a single
Nvidia 4090 GPU with 24GB memory. On the other hand, for image manipulation tasks, since different objects have varying visual appearances and thus different generation curves, it causes varying convergence speeds and thus different numbers of iterations during curve optimization.



\newpage
\appendix
\section{Curve Matching Algorithm}
\label{app:algorithm}
As described in Section~\ref{sec:curve_matching}, our curve matching algorithm manipulates the original image $X_0$ to align the generation curves of a source area $\mathcal{A}$ and a reference pixel $p^\star$.  
When applied for different image manipulation tasks, we flexibly select the reference pixel and modify the loss function $\mathcal{L}$ to achieve different goals. The default hyperparameter setting is $t=700$ and learning rate $\eta=0.02$. The iteration number ranges from 30 to 300 for different tasks, as we show their intermediate results during the optimization process. We use the pre-trained unconditional diffusion model, stable-diffusion-2-1-base, for all our experiments.

We present the details in Algorithm~\ref{alg}. The basic idea is to update $X_t$, transformed from the input $X_0$, in order to align the generation curves of the source area $c(X_t, \mathcal{A})$ and the curve $c^\star=c(X_t, p^\star)$ of the reference pixel $p^\star$ for each channel. Specifically, we first transform $X_0$ to $X_t$ via the deterministic process, and pre-compute the generation curve of the reference pixel and that of a random pixel within the source area, i.e. $c^\star =  r_t(X_t,v^\star)$, $c_a = r_t(X_t,v_a)$. Here we use the curve of a random point in $\mathcal{A}$ to represent the  $c(X_t, \mathcal{A})$. Then we start the iterative optimization. In each iteration, we sample a pixel $p_k$ within the source area randomly and a time $t_k$ based on the cumulative distribution of $c_a$, the generation curve of the representative source pixel. The optimization variable $X_t$ is transformed to $X_{t_k}$, which is used to compute the generation rate $r_{t_k}(X_{t_k}, v_k)$ and the loss function $\mathcal{L}  = |r_{t_k}(X_{t_k}, v_k) - r_{t_k}^\star|$. Then we update $X_t$ with an Adam optimizer.  After finishing the optimization, e.g. reaching the maximum iteration number, we recover $X_t$ back to $\bar{X}_0$ via Eq.~\ref{eq:ddim_ode}. 

\textbf{Curve updating.} 
The optimization objective requires to sample $t_k$ according to the curve of the source area, which in turn varies during the optimization and is time-consuming to compute. Firstly, to simplify the computation, we sample a representative pixel $p_a$ within the source area and compute its generation curve, i.e. $c_a$, to represent the curve of the source area. On the other hand, we update the curve after every $m=50$ iteration steps. Additionally, when the reference curve also varies, e.g. when the reference pixel lies within the source area, we update the reference curve as well.

\begin{algorithm}
\renewcommand{\algorithmicrequire}{\textbf{Input:}}
\renewcommand{\algorithmicensure}{\textbf{Output:}}
    \caption{Curve Matching Algorithm}
    \begin{algorithmic}[1]
    \REQUIRE Image $X_0$, source pixel set $\mathcal{A}$, reference pixel $p^\star$, hyperparameter $t$, iteration number $N$
    \Statex
    \Statex /* Initialization */
    \STATE Transform $X_0$ through Eq.~\ref{eq:ddim_ode} to obtain $X_{t}$ // Initialize the optimization variable
    \STATE Initialize directional vector $v^\star$ for $p^\star$ and $v_a$ for a randomly sampled source pixel $p_a \in \mathcal{A}$
    
    \Statex
    \Statex /* Pre-processing */
    \STATE $c_a\gets r_t(X_t,v_a)$, $c^\star\gets r_t(X_t,v^\star)$ // Compute curves $c_a$ and $c_r$ for $p_a$ and $p_r$ respectively
    
    \Statex
    \Statex /* Curve optimization with $X_{t}$ as the variable */
    \FOR {$k=1$ to $N$}
    \STATE Sample a random source pixel $p_k \in \mathcal{A}$ and initialize its corresponding vector $v_k$
    \STATE Sample a time $t_k$ based on the normalized $c_a$
    \STATE Transform $X_{t}$ to $X_{t_k}$ through Eq.~\ref{eq:ddim_ode} without gradients
    \STATE Compute generation rate $r_{t_k}(X_{t_k}, v_k)$
    \STATE Compute the loss $\mathcal{L}  = |r_{t_k}(X_{t_k}, v_k) - c^\star(t_{k})|$
    \STATE Update $X_{t}\gets X_{t}-\eta\nabla \mathcal{L} $
    \ENDFOR
    \Statex
    \Statex /* Generate the manipulated image from optimized $X_{t}$ */
    \STATE Transform the edited $X_{t}$ to $\hat{X}_0$ through Eq.~\ref{eq:ddim_ode}
    \end{algorithmic}
\label{alg}
\end{algorithm}
\section{Visual Analysis Experiments}
\label{sec:visual_analysis_appendix}

In the following visual analysis experiments, we take the $0$-th channel to represent a pixel for acceleration purpose. Although the generation curves of the channels at the same pixel have different values, we empirically found that they exhibit the same trend, allowing us to take one channel as the representative.

\subsection{Visual Saliency Experiment}
\label{app:saliency_val}
\textbf{Experiment setting.} We validate the connection between the fluctuation of our generation curve and the visual saliency of images, as described in Section~\ref{sec:analysis}. The experiment is conducted on the MIT saliency benchmark CAT2000~\cite{borji2015cat2000}, which provides the collected eye-tracking data of images from human observers and the pre-processed saliency map. For each image, we randomly select one pixel within maximum saliency values and one within minimum values as the salient pixel and non-salient pixel. For the generation curves at the two pixels, we compute its local variance for $t>200$ (to ignore the abrupt rise when $t$ is close to 0 ) over a sliding window of length $k=5$ and take their average to represent the curve fluctuation. 

\begin{figure}[h!]
  \centering
  \includegraphics[width=1\textwidth]{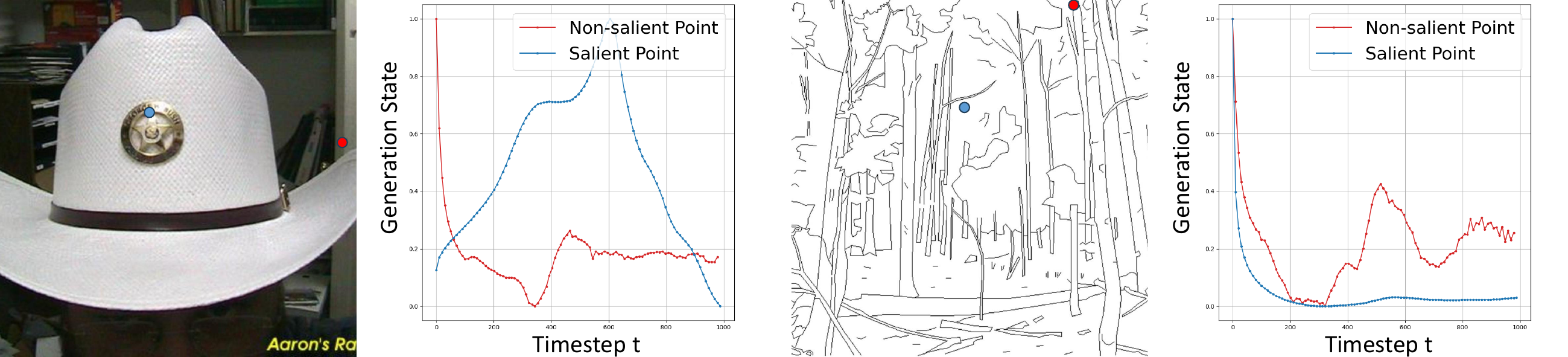} 
  \caption{The salient pixel (blue) and non-salient pixel (red) on the images and their corresponding generation curves. The left shows that our curve fluctuation is often consistent with the ground-truth saliency for natural images. The right is an inconsistent case with the special line drawing images.}
  \label{fig:more_saliency_val}
\end{figure}

\textbf{Discussion.} Figure~\ref{fig:more_saliency_val} presents the example images and the generation curves corresponding to the selected pixels. For the example on the left, the salient pixel corresponds to the blue curve with obviously higher fluctuation than the red curve. In our experiment, we found that our estimated curve fluctuation often reflects the visual saliency well for natural images. However, for some special cases, the noise inherent in eye-tracking data causes the inaccurate spatial location of the salient and non-salient pixels, and thus significantly interfering with our pixel-level calculations. For the line drawing images on the right, our curve fluctuation is not consistent with the ground-truth visual saliency. The red pixel is marked as non-salient pixel, while it corresponds to a higher curve fluctuation since it is located in the region with dense line drawings.

\subsection{Two types of Generation Curves}
\label{app:generation_val}
\textbf{Experiment setting.} Given an image and the mask of a small object, we compare our generation curve and another alternative approach, both of which estimate the generation rates of the small object at different timesteps. The alternative approach computes a perceptual-based curve that utilizes perceptual loss to estimate the generation rates of a local region. The experiments are as follows. 

\begin{itemize}
  \item \textbf{Perceptual-based curve.} For each timestep during the diffusion process, given the noised image $X_t$, we can predict the $\hat{X}_0$ with DDIM (Eq~\ref{eq:ddim_ode}) and decode the corresponding RGB image $\hat{I}$. Then we use the pre-trained DINO model~\cite{caron2021emerging} to compute the perceptual loss between the predicted $\hat{I}$ and the original image $I$ w.r.t. the region of the specified small object. The perceptual loss is defined as the cosine distance between the features $DINO(\hat{I}\odot M)$ and $DINO(I\odot M)$, corresponding to the generation state at timestep $t$ that ranges from $0$ to $1$. And its derivatives can be considered as an approximation of the generation rates. 
  \item \textbf{Our generation curve.} Due to the compressive nature of LDM and the sensitivity of our method at the pixel-channel level, we select the central pixel of the specified small object as the representative spatial location to compute our generation curve. That is, given the noised image $X_t$, we construct the vector $v$ based on the pixel location and compute the generation rates $r_t(X_t,v)$ with Eq~\ref{eq:curve_computation} to obtain the generation curve.
\end{itemize}

\begin{figure}[h!]
  \centering
  \includegraphics[width=1\textwidth]{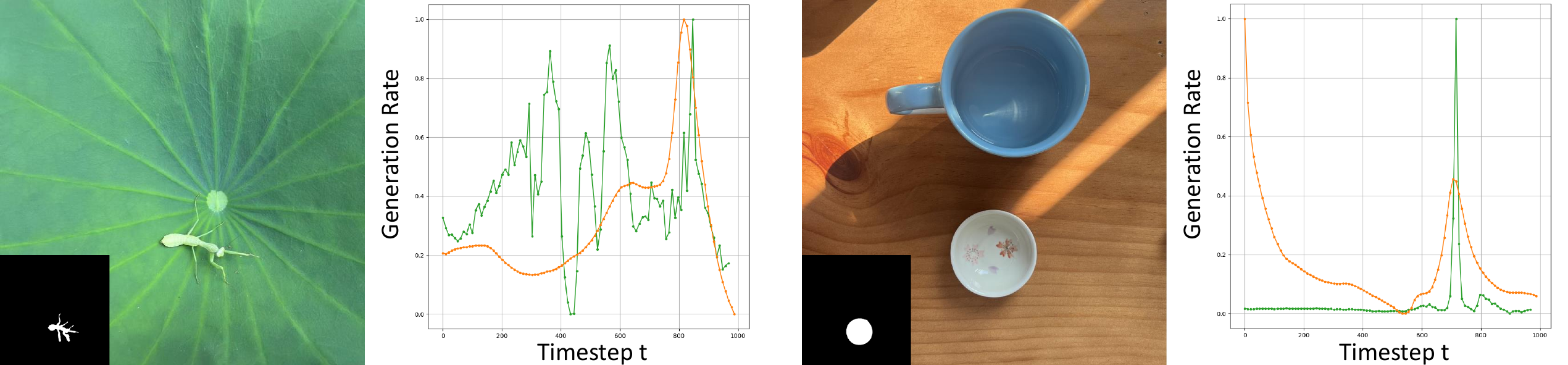} 
  \caption{Two types of generation curves for the masked small object in the image. Green: the perceptual-based curve; Orange: our generation curve. All the curves are normalized into range $[0,1]$.}
  \label{fig:more_generation_val}
\end{figure}

\begin{figure}[h!]
  \centering
  \includegraphics[width=0.5\textwidth]{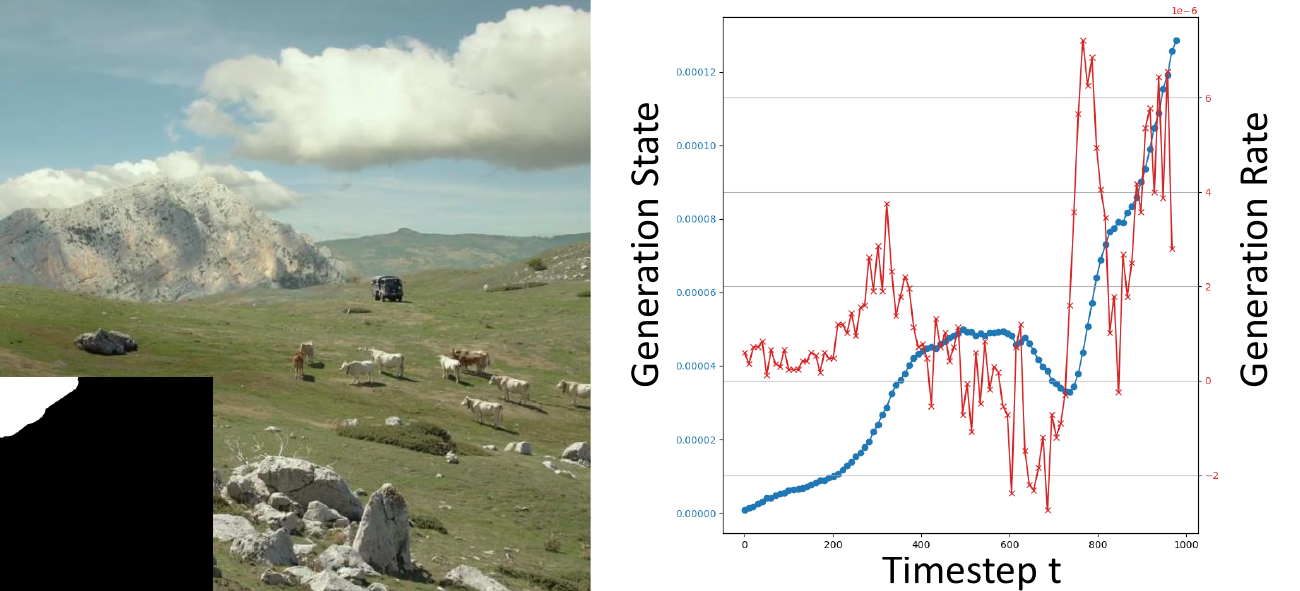} 
  \caption{We plot the generation state (blue) and generation rate (red) estimated by the perceptual-based loss. For background areas without prominent visual features, the perceptual-based curves (red) tend to exhibit heavy noise or even negative generation rates.}
  \label{fig:generation_failure}
\end{figure}

\textbf{Discussion.} Figure~\ref{fig:more_generation_val} presents the curves of two example images, both ours and the perceptual-based curves. Note that we normalize the curves into the range $[0,1]$ for a better visualization. As described in Section~\ref{sec:analysis}, the two types of curves exhibit similar trends, especially with their main peaks occurring at close timesteps. It validates that both the two curves reflect the generation rates. 
On the other hand, compared to the intuitive perceptual-based curves, our generation curve is more applicable to many potential tasks.
Firstly, the perceptual loss is object-wise rather than pixel-wise. It prevents its analysis and application to more fine-grained generation patterns. 
Secondly, the perceptual loss can only distinguish the prominent foreground features. Consequently, the perceptual-based curve often exhibits heavy noise or even negative values with less prominent visual features, especially the flat background, as shown in Figure~\ref{fig:generation_failure}.

\section{Application Details}
\label{app:application}

\subsection{Semantic Transfer}

\begin{figure}[!ht]
  \centering
  \includegraphics[width=1\textwidth]{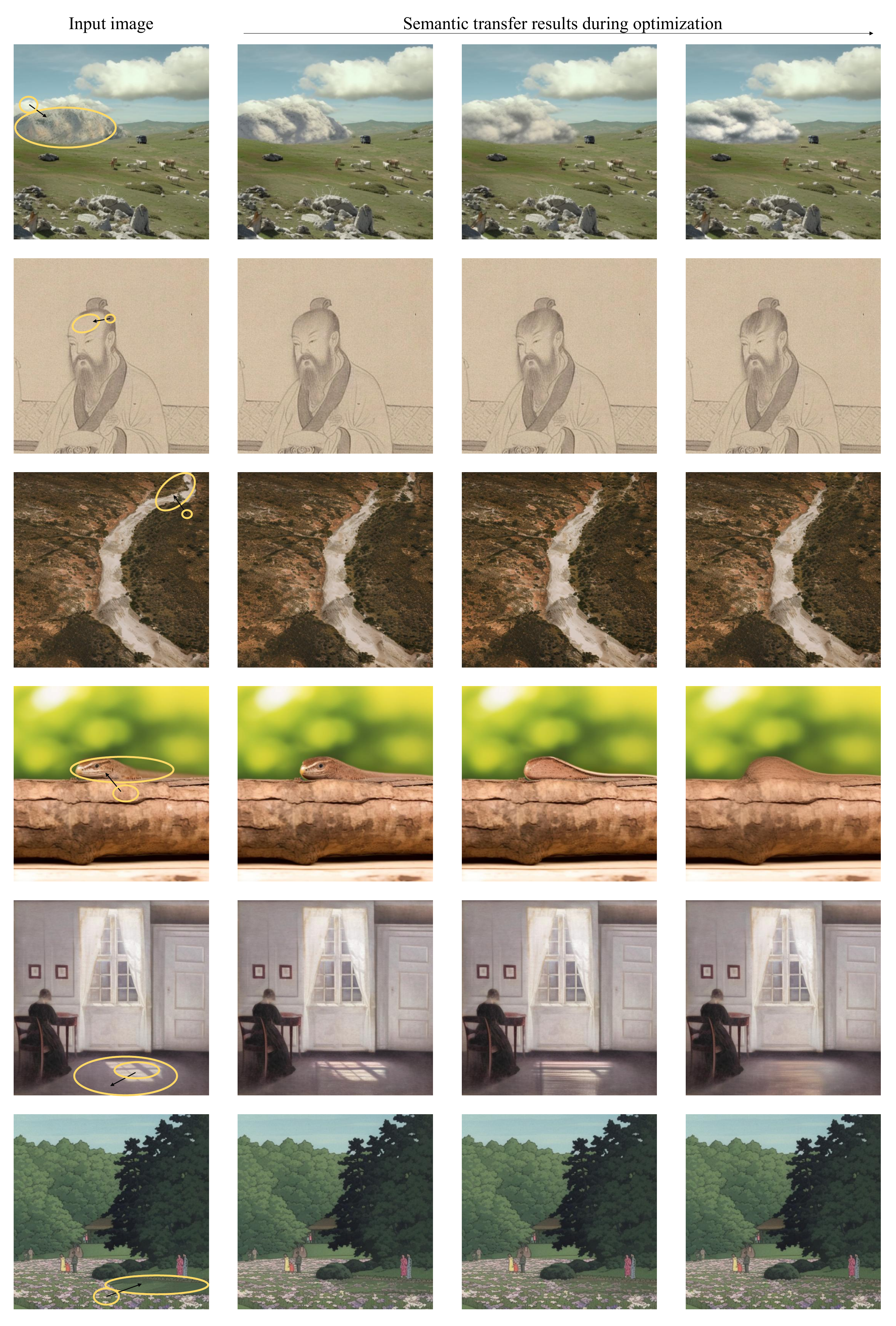} 
  \caption{More results of semantic transfer. The left column shows the input image, where the arrows indicate the semantic transfer from the reference to source area. From left to right, we show the intermediate results during the optimization.}
  \label{fig:more_res_transfer}
\end{figure}

\begin{figure}[ht!]
  \centering
  \includegraphics[width=1\textwidth]{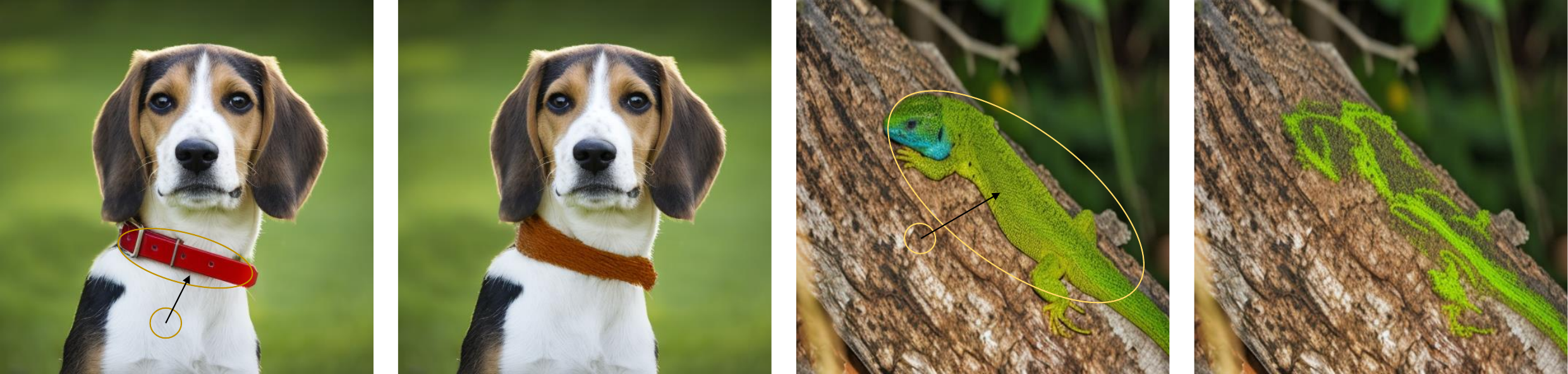} 
  \caption{Failure cases of semantic transfer. The material and depth features are transferred but not the color.}
  \label{fig:fail_res_transfer}
\end{figure}

The semantic transfer application shares the same pipeline with our object removal application, except that we select a representative pixel of a surrounding object as the reference. We demonstrate more visual results in Figure~\ref{fig:more_res_transfer}. It enables interactive editing with only a click on the image to specify the reference. However, since the generation curves involve various visual properties, sometimes the transfer results don't exhibit the desired visual effects. For example, as shown in Figure~\ref{fig:fail_res_transfer}, it transfers the material but not the color in the two examples.

\subsection{Object Removal}

\textbf{Experiment setting.} We conduct the comparison experiment on the test set from Emu Edit benchmark~\cite{sheynin2023emu}. For the object removal task, it provides the original images, the input and output captions, and the text instructions to specify the objects to be removed. We pre-process the test set with an image segmentation tool to obtain the masks of the objects to be removed. The quantitative evaluation is performed on a random subset of 100 images from this test set. Note that there's no need for the training set since we utilize the pre-trained models in all our experiments.

\begin{figure}[h!]
  \centering
  \includegraphics[width=1\textwidth]{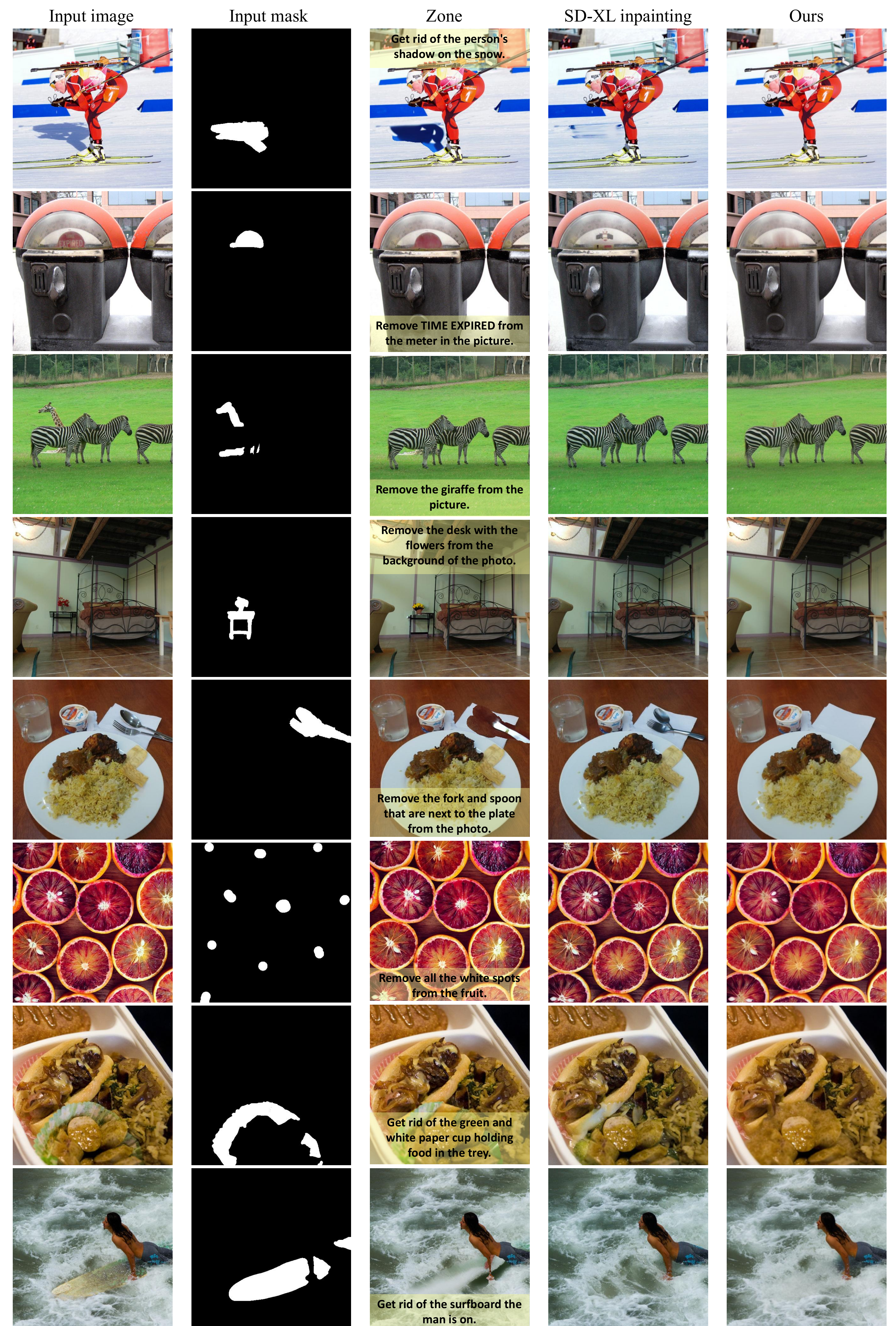} 
  \caption{More object removal results. From left to right: the input image and the input object mask, the results of the instruction-based method ZONE~\cite{li2024zone}, SD-XL inpainting~\cite{podell2023sdxl}, and our approach.}
  \label{fig:more_res_removal}
\end{figure}

The related works often utilize image inpainting~\cite{Criminisi2004RegionFA, suvorov2021resolutionrobust, lugmayr2022repaint, podell2023sdxl} or instruction-based image editing~\cite{brooks2023instructpix2pix, pan2023drag, hertz2022prompttoprompt} for the object removal task. The former resamples the editing area, which is then likely to fall into the high-density distributions, i.e. image background. The latter fine-tunes the pre-trained image generation model to take the text instructions as conditions. We compare with the SD-XL inpainting~\cite{podell2023sdxl} and instruction-based method ZONE~\cite{li2024zone} in our experiments:
\begin{itemize}
  \item \textbf{SD-XL inpainting.} We input the original image and the object mask to the SD-XL model, i.e. stable-diffusion-xl-1.0-inpainting-0.1, and obtain the inpainted image as the result.
  \item \textbf{ZONE.} We take the original image and the text instruction as input and the generated image as result. Note that the pre-processed object mask is used in the composition of the generated image for a precise local editing.
  \item \textbf{Ours.} The input includes the original image, the object mask as the source area, a selected surrounding background pixel as the reference. Note that the tedious selection of reference pixels can be omitted for simple cases with a uniform background, as described in the following implementation.
\end{itemize}

\textbf{Implementation.} To alleviate the manual selection of reference pixels, we adopt an approximation solution for the simple cases. That is, we divide the test set into two types: one containing the target objects located on uniform backgrounds, and the other with complex and varying backgrounds. For the former, instead of using the generation curve at a manually selected reference pixel, we define a fixed pseudo curve to replace it. Specifically, since the generation curves at background locations often follow the common pattern with stable and lower values, we directly minimize the generation rate values at the target area for $t>200$, i.e. $\mathcal{L}  = |r(X_{t_i}, p_i)|$. For the latter, we manually select the reference pixels and invoke our curve matching algorithm as described in Algorithm~\ref{alg}. We experimentally found that the pseudo curve solution is enough for most images (more than 80 out of the 100 images in our quantitative evaluation, Table~\ref{tab:removal_quantitative1} in the main paper).

\begin{figure}[t]
  \centering
  \includegraphics[width=1\textwidth]{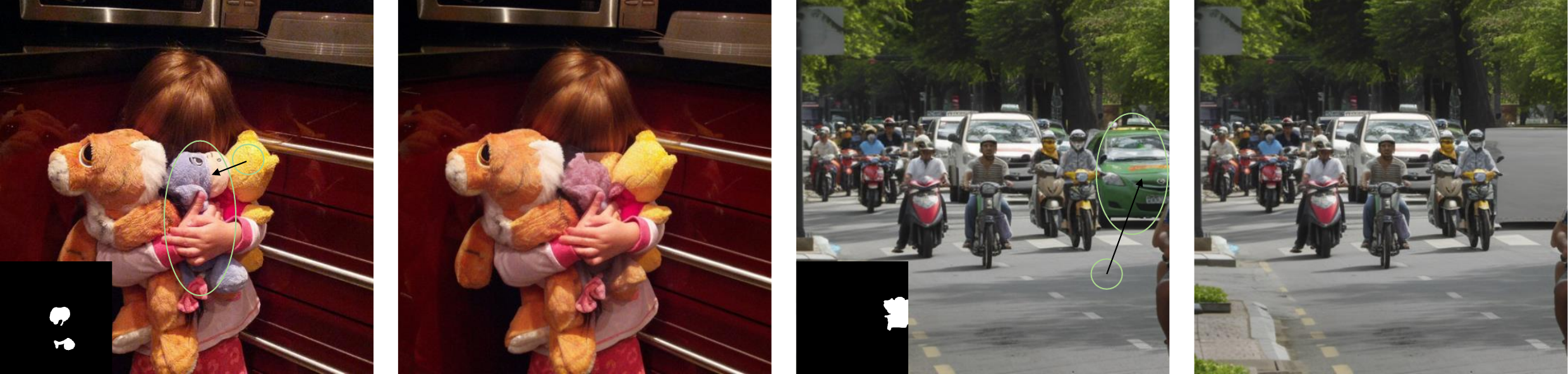} 
  \caption{Failure cases of object removal application.}
  \label{fig:fail_res_removal}
\end{figure}

\textbf{Results and failure cases.} We present more results of object removal in Figure~\ref{fig:more_res_removal}. Our approach outperforms the alternatives in terms of generating clean and reasonable results. However, we also notice some failure cases when the background is complex. As shown in Figure~\ref{fig:fail_res_removal}, sometimes it may be hard to specify a reference pixel representative for the background. This often causes distorted results at the region of the object to be removed.  

\subsection{Saliency Manipulation}

\begin{figure}[ht!]
  \centering
  \includegraphics[width=1\textwidth]{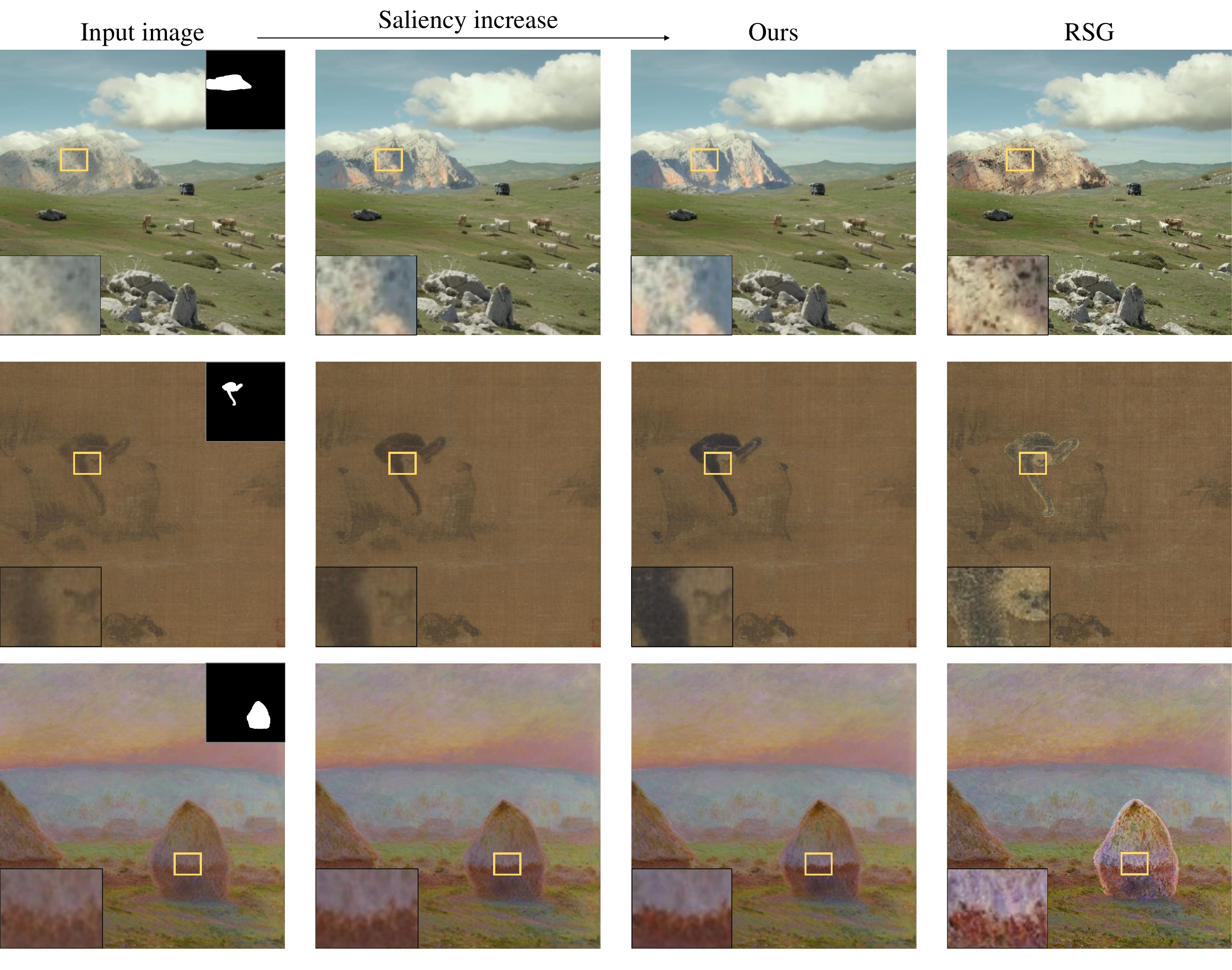} 
  \caption{Saliency increasing results. From left to right: the input image, the intermediate and final results of our approach, the result of RSG~\cite{miangoleh2023realistic}.}
  \label{fig:more_res_saliency_increase}
\end{figure}
\begin{figure}[t]
  \centering
  \includegraphics[width=1\textwidth]{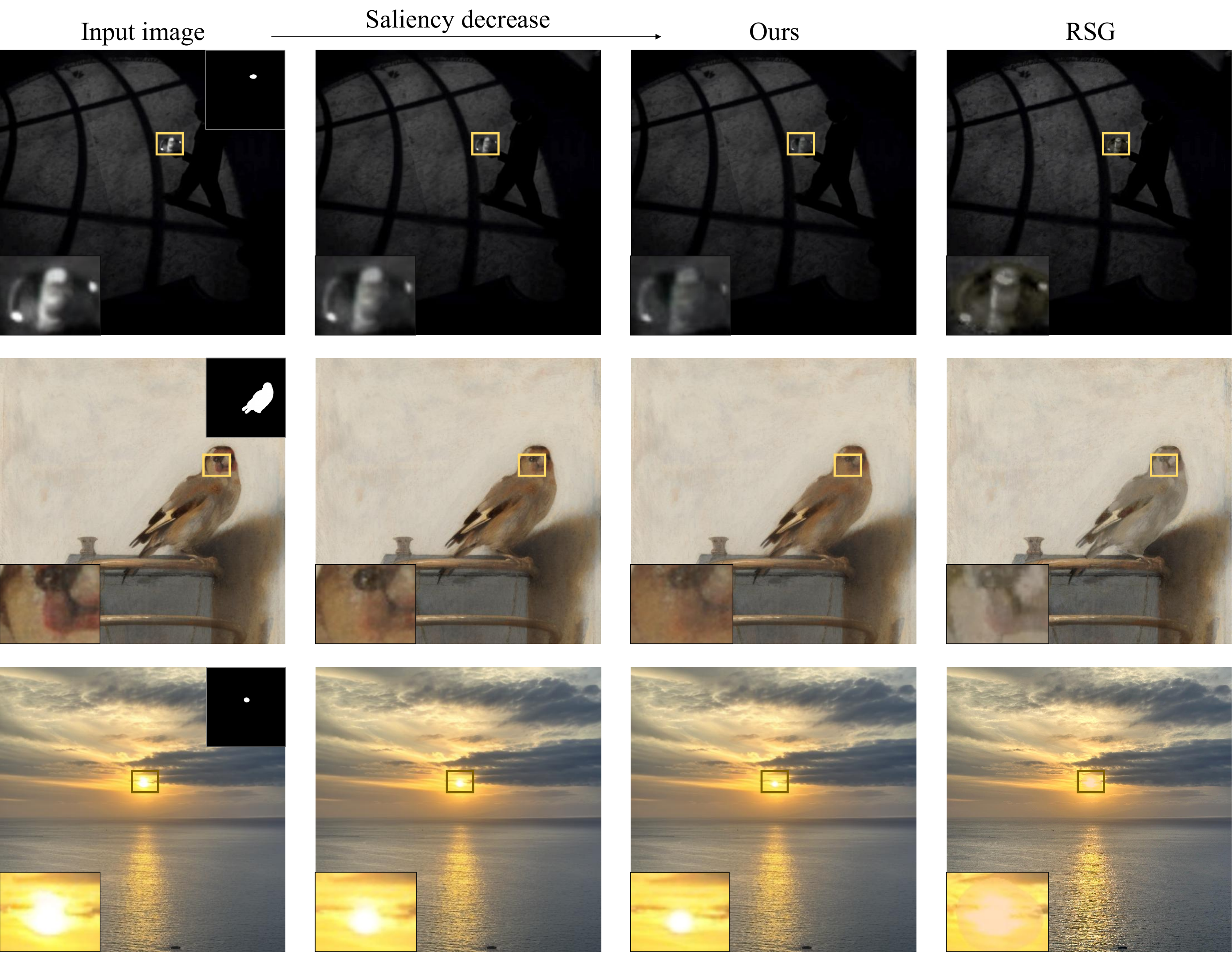} 
  \caption{Saliency decreasing results. From left to right: the input image, the intermediate and final results of our approach, the result of RSG~\cite{miangoleh2023realistic}.}
  \label{fig:more_res_saliency_decrease}
\end{figure}

\begin{figure}[h!]
  \centering
  \includegraphics[width=1\textwidth]{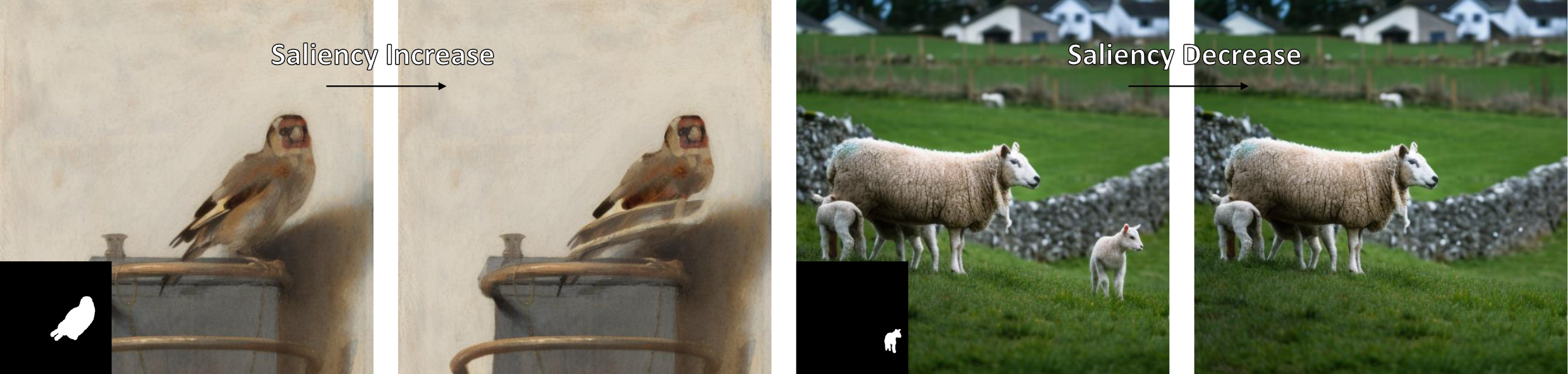} 
  \caption{Failure cases of saliency manipulation.}
  \label{fig:fail_res_saliency_decrease}
\end{figure}

\textbf{Experiment setting.} Saliency manipulation refers to increase or decrease the saliency of a specific object as one expects, while maintaining the image content as unchanged as possible~\cite{Aberman2021DeepSP,Mejjati2020LookHA,Jiang2021SaliencyGuidedIT,miangoleh2023realistic}. The input contains the original image and a mask indicating the region to be edited. In our experiments, we compare against a recent saliency-based image manipulation approach~\cite{miangoleh2023realistic}. 
This approach, denoted as RSG, optimizes the image with a saliency loss using a pre-trained saliency model and a realism loss to prevent frequent unrealistic edits. 
We use their released code and the parameters in our experiments. As for our approach, we eliminate the reference curve, but directly minimize or maximize the generation rate values for $t>200$ at the editing region to control the saliency. At the same time, we define a feature alignment loss to preserve the original image content. The feature alignment loss is defined as the difference between the U-Net intermediate feature maps before and after the editing. In summary, we replace the loss in Algorithm~\ref{alg} as $\mathcal{L}  = \lambda _1 |r_{t_k}(X_{t_k}, p_k)|^{\lambda _2}  + \sum_i |U_i(X_{t_k}) - U_i(X_t)|$, where $\lambda _1=50$ is a weighting parameter and $\lambda 2$ takes values $1$ or $-1$ to decrease or increase the saliency, respectively. 
And $U_i$ represents the U-Net layers which outputs the intermediate feature maps. We set the iteration number $N=70$ for the saliency manipulation task. 

\textbf{Results and failure cases.} We present the saliency increase and decrease examples in Figure~\ref{fig:more_res_saliency_increase} and \ref{fig:more_res_saliency_decrease}. As the iteration number increases during the optimization, our approach obtains edited objects with saliency varies as expected. Increasing the saliency results in sharp color contrast and more visual details, while decreasing the saliency results in faded visual effects. We also present the failure cases in Figure~\ref{fig:fail_res_saliency_decrease}. It is worth noting that all the saliency manipulation approaches require a balance between local editing and the preservation of image content. Sometimes, forcing the saliency variation may cause the altering of the original identity and object distortion.

\subsection{Image Blending}

\textbf{Experiment setting.} Image blending aims to create a natural boundary transition for the compositional images~\cite{Prez2003PoissonIE, wu2019gpgan, zhang2019deep, zhang2020deep, Xing2022CompositePH,niu2024making}. We evaluate the image blending results on the iHarmony4 dataset~\cite{DoveNet2020}. It provides the synthesized composite image with inconsistent foreground and background, as well as the corresponding foreground masks. 
To conduct the experiment with our approach, we apply the erosion filter with kernel size $k=3$ on the given mask and take the eroded region as the source area of our curve matching algorithm. We use the algorithm with loss $\mathcal{L}  = \lambda _1|r_{t_k}(X_{t_k}, p_k)| + \sum_i |U_i(X_{t_k}) - U_i(X_t)|$ to complete this task, where $\lambda _1=50$ is a weighting parameter and $p_i$ are the pixels within the eroded region of the mask. We set the iteration number $N=100$ for the image blending task. 

\begin{figure}[t]
  \centering
  \includegraphics[width=1\textwidth]{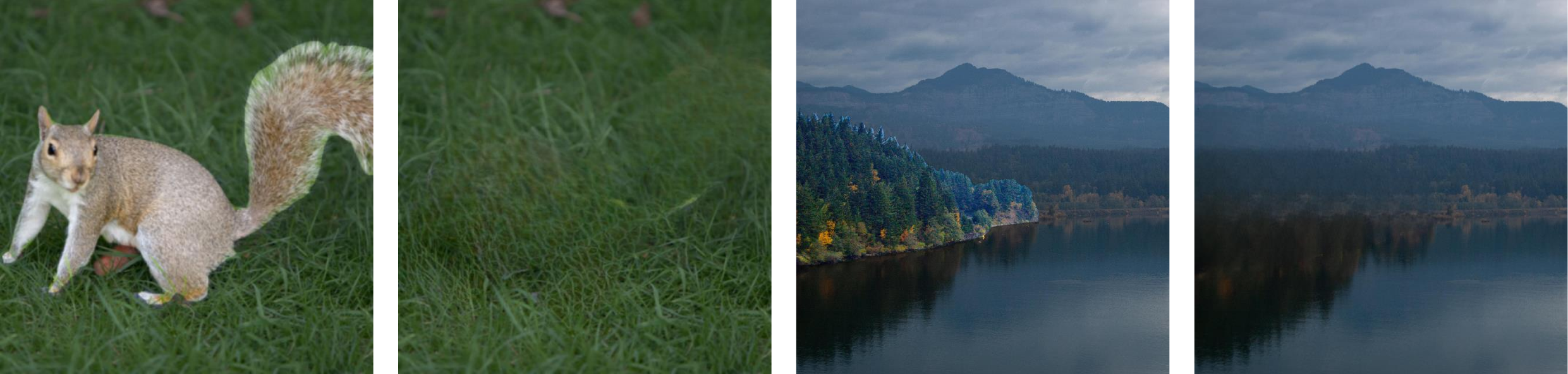} 
  \caption{Provided background for GP-GAN~\cite{wu2019gpgan} in image blending.}
  \label{fig:blending_bg}
\end{figure}

\begin{figure}[t]
  \centering
  \includegraphics[width=1\textwidth]{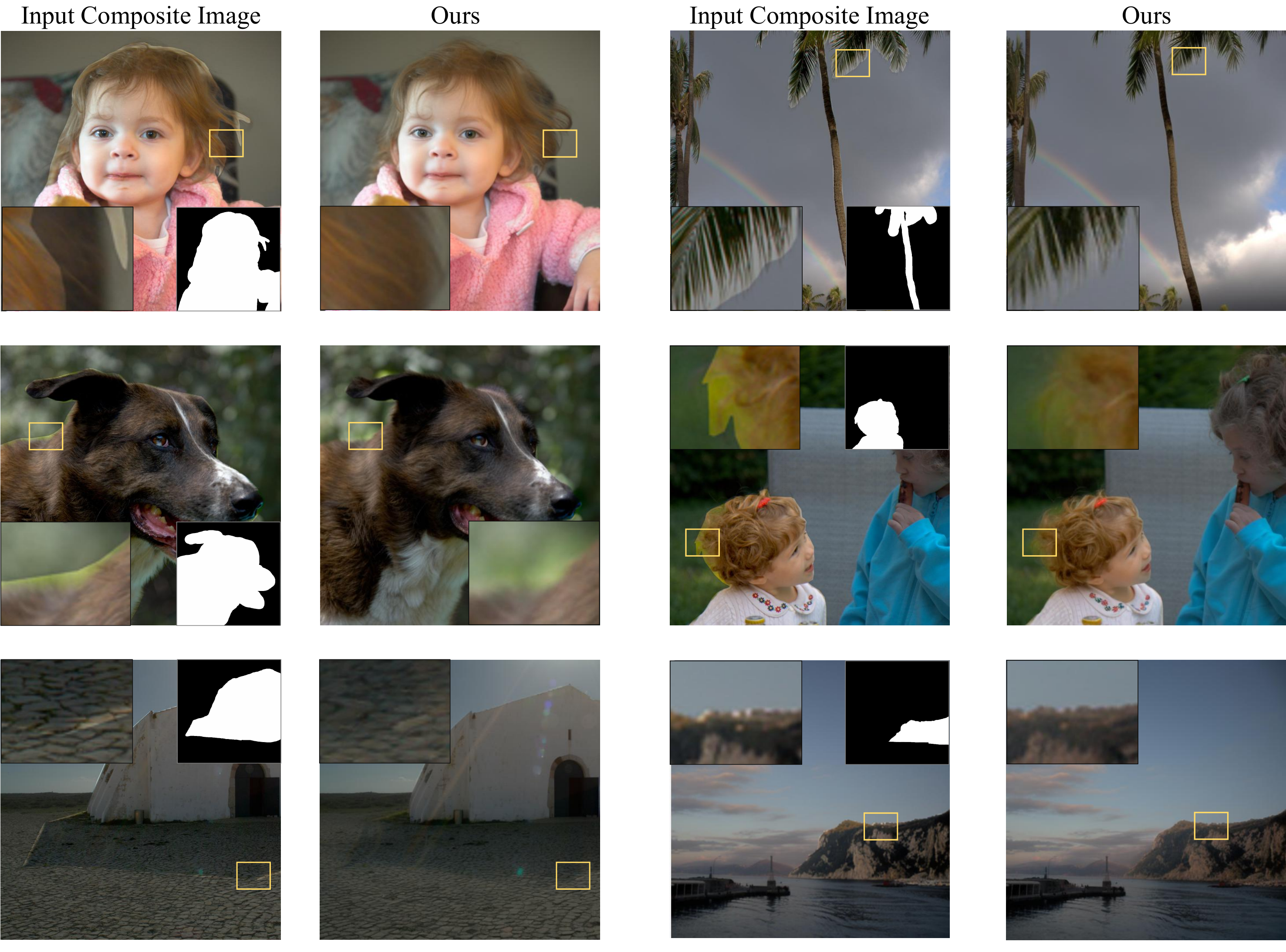} 
  \caption{Image blending results. For each example, we show the input composite image and the foreground mask, as well as the result of our approach.}
  \label{fig:more_res_blending}
\end{figure}

\begin{figure}[h!]
  \centering
  \includegraphics[width=1\textwidth]{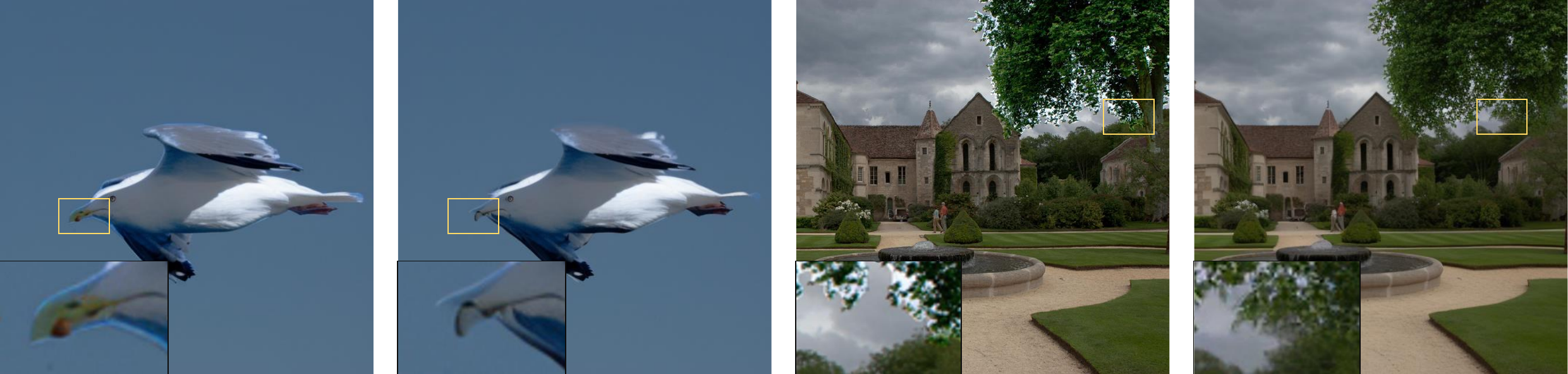} 
  \caption{Failure cases of image blending. Although we achieve a natural blending at the boundary, it loses the small details such as the beak of the bird and the leaves on the tree.}
  \label{fig:fail_res_blending}
\end{figure}

\textbf{Results and failure cases.} 
When compared to~\cite{wu2019gpgan} in Figure~\ref{fig:blending_visual}, this approach requires both the complete background and foreground images as input, Therefore, we utilize online tools that combines the inpainting~\cite{suvorov2021resolutionrobust} and post-processing techniques to provide a suitable background, as shown in Figure~\ref{fig:blending_bg}. We present more results in Figure~\ref{fig:more_res_blending}. Our approach is able to produce a natural and smooth transition at the boundary. However, the smooth transition often corresponds to blurred details. As a consequence, the image details might be smoothed out during the blending, such as the beak of the bird and the leaves on the tree in Figure~\ref{fig:fail_res_blending}.

\section{Broader Impact} Similar to other image manipulation technologies, our generation curve matching algorithm also suffer the potential ethical implications such as disinformation or generating fake images. We recognize the potential ethical concerns that could emerge from utilizing our approach. We strongly encourage the creation and application of social and technical safeguards to mitigate potential misuse, such as the dissemination of disinformation or propaganda. We are dedicated to upholding fairness and non-discrimination, legal adherence, and research integrity in our endeavors.


\end{document}